\documentclass[twoside,11pt]{article}
\usepackage{pgm}
\usepackage{amsmath}
\usepackage{latexsym}
\usepackage{subcaption}
\usepackage{algorithm}
\usepackage[dvipsnames]{xcolor}
\usepackage[noend]{algpseudocode}
\usepackage{tikz}
\usetikzlibrary{bayesnet}
\usetikzlibrary{positioning}

\usepackage[utf8]{inputenc}
\inputencoding{utf8}

\usepackage{hyperref,color,soul,booktabs}
\setulcolor{blue}
\newcommand{\BibTeX}{\textsc{B\kern-0.1emi\kern-0.017emb}\kern-0.15em\TeX}

\newcommand\shrink[1]{}

\def\n(#1){\bar{#1}}

\def\data{\mathcal{D}}
\def\pr{{\it Pr}}

\def\d{{\bf d}}

\def\e{{\bf e}}

\def\U{{\bf U}}
\def\u{{\bf u}}

\def\v{{\bf v}}

\def\X{{\bf X}}
\def\x{{\bf x}}
\def\Y{{\bf Y}}
\def\y{{\bf y}}

\def\eql(#1,#2){{#1\!\!=\!#2}}

\def\eql(#1,#2){{#1\!=\!#2}}
\newcommand\name[1]{\ensuremath{\mathsf{#1}}}
\def\true{\name{true}}
\def\false{\name{false}}

\newcommand{\argmax}{\operatornamewithlimits{arg max}}

\def\clap#1{\hbox to 0pt{\hss#1\hss}}

\newcommand\Amain[1]{
\textbf{main:}
#1
\vspace{1mm}
}

\newcommand\Ainputb[4]{
\vspace{1mm}
\textbf{input:}\\
\setlength{\tabcolsep}{4pt}
\renewcommand{\arraystretch}{1}
\begin{tabular}{ll}
#1: & #2 \\
#3: & #4
\end{tabular}
}

\newcommand\Aoutput[1]{
\vspace{1mm}
\textbf{output:} #1
\vspace{1mm}
}

\def\funky{{FoCS}}
\def\idthree{\ensuremath{\mathsf{ID3}}}
\def\learncontext{\ensuremath{\mathsf{LearnContext}}}

\begin{document}

\title{A New Perspective on Learning Context-Specific Independence}

\author{\Name{Yujia Shen} \Email{yujias@cs.ucla.edu}\and
  \Name{Arthur Choi} \Email{aychoi@cs.ucla.edu}\and
  \Name{Adnan Darwiche} \Email{darwiche@cs.ucla.edu}\\
  \addr Computer Science Department, University of California, Los Angeles}

\maketitle

\begin{abstract}%   <- trailing '%' for backward compatibility of .sty file
Local structure such as context-specific independence (CSI) has received much attention in the probabilistic graphical model (PGM) literature, as it facilitates the modeling of large complex systems, as well as for reasoning with them. In this paper, we provide a new perspective on how to learn CSIs from data. We propose to first learn a functional and parameterized representation of a conditional probability table (CPT), such as a neural network. Next, we quantize this continuous function, into an arithmetic circuit representation that facilitates efficient inference. In the first step, we can leverage the many powerful tools that have been developed in the machine learning literature. In the second step, we exploit more recently-developed analytic tools from explainable AI, for the purposes of learning CSIs. Finally, we contrast our approach, empirically and conceptually, with more traditional variable-splitting approaches, that search for CSIs more explicitly.
\end{abstract}
\begin{keywords}
context-specific independence,
Bayesian networks,
arithmetic circuits.
\end{keywords}

\newcommand{\jason}[1]{\textcolor{red}{[Jason: #1]}}

\section{Introduction}

Context-specific independence (CSI) is a type of local structure that facilitates the modeling of large and complex systems, by allowing one to represent in a succinct way conditional distributions that would otherwise be infeasible to represent \cite{BoutilierFGK96}.  Further, local structure such as context-specific independence can be exploited by modern classes of inference algorithms to perform reasoning in Bayesian networks whose treewidths are too large for more traditional inference algorithms \cite{darwicheJACM-POLY,Chavira.Darwiche.Aij.2008,ShenCD16}.
Traditional representations of context-specific independence (CSI) use data structures such as decision trees, decision graphs, rules, default tables, etc. \cite{friedman1998learning,chickeringHM97,LarkinD03,KollerFriedman}.  Algorithms for learning these representations are typically search-based, where we iteratively search for variables that split the data into partitions, until the resulting distribution becomes (sufficiently) independent of the remaining variables.

In this paper, we propose a new perspective on learning CSIs, resulting in a new context-specific representation for conditioanl probability tables (CPTs) that we call \underline{F}uncti\underline{o}nal \underline{C}ontext-\underline{S}pecific CPTs, or just \funky\ CPTs.  \funky\ CPTs generalize rule CPTs, where a context is typically defined as a partial instantiation of the variables.  More recently, the Conditional Probabilistic Sentential Decision Diagram (or Conditional PSDD) was proposed, and further generalizes term-based rules to arbitrary propositional sentences \citep{ShenCD18}.  \funky\ CPTs generalize this further so that an arbitrary function can be used to define the scope of a context, say one defined by a neural network.

The first significance of this new representation is that it allows us to immediately leverage powerful machine learning systems that have been developed in recent years, for the purposes of learning CSIs.  
%Unfortunately, as we show later, functional CPTs do not allow for efficient probabilistic inference.
%
The second significance is that efficient probabilistic can become enabled, by exploiting recently developed analytic tools from the domain of eXplainable Artificial Intelligence (XAI),\footnote{\url{https://www.darpa.mil/program/explainable-artificial-intelligence}.} which allows us to extract a decision graph representation of a context-specific CPT, but one that facilitates exact inference, i.e., a conditional PSDD \cite{ShenCD18,ShenGCD19,ShenCD16}.

This paper is organized as follows.  In Section~\ref{sec:review}, we review functional and context-specific representations of CPTs.  In Section~\ref{sec:model} we propose the \funky\ CPT.  In Section~\ref{sec:learn} we propose an algorithm to learn \funky\ CPTs from data, and in Section~\ref{sec:reason} we show how to reason with them.  We empirically compare \funky\ CPTs with functional and context-specific representations in Section~\ref{sec:experiments}, and we provide a case study on ``learning to decode'' in Section~\ref{sec:case-study}.  Finally, we conclude in Section~\ref{sec:conclusion}.

\section{Representations of CPTs} \label{sec:review}

A Bayesian network (BN) has two main components: (1) a directed acyclic graph (DAG) and (2) a set of conditional probability tables (CPTs) \cite{Pearl88b,Darwiche09,KollerFriedman,MurphyBook}.  Typically, CPTs are represented using tabular data structures, although this becomes impractical when a variable has many parents.  In this section, we review two alternative representations of interest: functional representations (such as noisy-or models and neural networks) and context-specific representations, such as tree CPTs and rule CPTs.

In what follows, we use upper case letters (\(X\)) to denote variables and lower case letters \((x)\) to denote their values. Variable sets are denoted by bold-face upper case letters (\(\X\)) and their instantiations by bold-face lower case letters (\(\x\)).  Generally, we use \(X\) to denote a variable in a Bayesian network and \(\U\) to denote its parents.  We further refer to \(X\U\) as a family.  We thus denote a network parameter using the form \(\theta_{x|\u},\) which represents the conditional probability \(\pr(\eql(X,x)|\eql(\U,\u))\).

\subsection{Functional Representations} \label{sec:functional-cpts}

To specify a CPT using a table, one must specify a parameter \(\theta_{x|\u}\) for all family instantiations \(x\u\), the number of which is exponential in the number of variables in the family \(X\U\).  In a functional representation of a CPT, one has a parametrized function \(f(X\U; \theta)\) that \emph{computes} the probability \(\pr(x|\u)\) from a parameter vector \(\theta\) that can be much smaller than the size of an explicit table.  For example, the well-known noisy-or model implicitly specifies a conditional distribution using a number of parameters that is only linear in the number of parents \cite{Pearl88b,Darwiche09}.

Other functional representations include logistic functions \citep{Frey98,Vomlel06} as well as neural networks \citep{Bengio2000,KingmaW13}.  Consider the following conditional distribution for a variable \(X\) with parents \(U_1 U_2\), where each variable is binary (\(0/1\)):
\(
\pr(\eql(x,1) \mid u_1 u_2) = \sigma(\theta_0 + \theta_1 u_1 + \theta_2 u_2),
\)
where \(\sigma(a) = [1+\exp\{-a\}]^{-1}\) is the sigmoid function and where \(\theta_0, \theta_1, \theta_2\) are parameters.  This functional CPT has the following tabular representation:
\begin{center}
\small
  %% \begin{tabular}{cc|rr}
  %%   \(u_1\) & \(u_2\) & \(\pr(\eql(X,1) \mid u_1 u_2)\)  & \(\pr(\eql(X,0) \mid u_1 u_2)\) \\ \hline
  %%   \(1\) & \(1\) & \(\sigma(\theta_0 + \theta_1 + \theta_2)\) & \(1-\sigma(\theta_0 + \theta_1 + \theta_2)\) \\
  %%   \(1\) & \(0\) & \(\sigma(\theta_0 + \theta_1)\) & \(1-\sigma(\theta_0 + \theta_1)\) \\
  %%   \(0\) & \(1\) & \(\sigma(\theta_0 + \theta_2)\) & \(1-\sigma(\theta_0 + \theta_2)\) \\
  %%   \(0\) & \(0\) & \(\sigma(\theta_0)\) & \(1-\sigma(\theta_0)\)
  %% \end{tabular}
  \begin{tabular}{c|cccc}
    \(u_1, u_2\) & \(1,1\) & \(1,0\) & \(0,1\) & \(0,0\) \\ \hline
\(\pr(\eql(X,1) \mid u_1 u_2)\) & \(\sigma(\theta_0 + \theta_1 + \theta_2)\)   & \(\sigma(\theta_0 + \theta_1)\)   & \(\sigma(\theta_0 + \theta_2)\)   & \(\sigma(\theta_0)\) \\
\(\pr(\eql(X,0) \mid u_1 u_2)\) & \(1-\sigma(\theta_0 + \theta_1 + \theta_2)\) & \(1-\sigma(\theta_0 + \theta_1)\) & \(1-\sigma(\theta_0 + \theta_2)\) & \(1-\sigma(\theta_0)\)
  \end{tabular}
\end{center}
In general, if we have \(n\) binary parents \(U\), then the tabular representation will have \(2^n\) free parameters, whereas the functional logistic representation will have only \(n+1\) parameters.

While functional representations allow us to compactly specify a CPT, they become unwieldly once we need to perform any reasoning.  For example, the following result shows that computing the Most Probable Explanation (MPE) is intractable when using a logistic representation of a CPT, even when the parents are independent.
The proof follows by reduction from the knapsack problem. % (a full proof is omitted for space).
\begin{theorem} \label{thm:difficult-structured-prediction}
Consider a prior probability \(\pr(\U)\) over \(n\) variables and a conditional probability \(\pr(X \mid \U).\) If the prior probability is fully factorized and \(\pr(X \mid \U)\) is represented by a logistic function using \(n+1\) parameters, it is NP-complete to compute \(\argmax_\u \pr(\u \mid x).\)
\end{theorem}

\subsection{Context-Specific CPTs}

A \emph{decision-tree CPT}, or just \emph{tree CPT}, represents a conditional distrubtion of a variable \(X\) given its parents \(\U\) in a Bayesian network \citep{friedman1998learning,desJardins05}.  It is composed of a decision tree over variables \(\U\), and at each leaf of the decision tree is a CPT column, which we denote by \(\Theta_{X|.}\).  A \emph{CPT column} \(\Theta_{X|.}\) is a distribution over variable \(X\) for some given context.  A \emph{decision-graph CPT} is a representation like a decision tree, but where equivalent leaves with equivalent CPT columns \(\Theta_{X|.}\) are merged together in a single node \cite{chickeringHM97}.  This decision graph can be further simplified by iteratively merging decision nodes whose children are equivalent.
A \emph{rule CPT} is another representation of a conditional distribution that uses rules to define the CPT.  A rule is composed of two parts: a context, which is typically a partial instantiation \(\v\) of the parent variables \(\U\), and a CPT column \(\Theta_{X|.}\).  A set of rules specificy a rule CPT if the contexts \(\v\) represent a mutually-exclusive and exhaustive partitioning of the instantiations of \(\U\).

A decision-tree CPT specifies a set of rules, where each leaf represents a rule with the same CPT column \(\Theta_{X|.}\) assigned to the leaf, where the context \(\v\) is found by taking the value of each variable that was branched on, on the path from the root to the leaf.  A decision-graph CPT also specifies a set of rules in a similar way, except that we relax the requirement that the context be specified by a partial instantiation, but now as a disjunction of partial instantiations, one for each path that can reach the leaf from the root.
The rule CPT can be generalized further by allowing the context to be specified as an arbitrary propositional sentence.  Such a rule CPT can be realized using the recently proposed Conditional Probabilistic Sentential Decision Diagram (Conditional PSDD) \citep{ShenCD18}.\footnote{In a Conditional PSDD, the contexts are represented using a shared SDD \cite{Darwiche11}, and the CPT columns are represented using a shared PSDD \cite{KisaVCD14}.}  While they enable more succinct representations of conditional distributions, Conditional PSDDs also facilitate the ability to reason with them \citep{ShenCD16,ShenGCD19}.

%\art{more examples here if we end up having space?}

\section{Functional Context-Specific CPTs} \label{sec:model}

Next, we propose a generalized rule CPT where the scope of a rule is defined, not just by a partial instantiation of its parents, or just by a propositional sentence, but more generally by some function.

\begin{definition}%[Functional Context-Specific CPT]
A \underline{F}uncti\underline{o}nal \underline{C}ontext-\underline{S}pecific (\funky) CPT represents a conditional distribution of a variable \(X\) given its parents \(\U\), and is defined by a set of \(k\) context/CPT-column pairs \((\Delta_i,\Theta_{X|\Delta_i})\) where the contexts \(\Delta_i\) are a mutually-exclusive and exhaustive partition of the instantiations of \(\U\).
\end{definition}
A \funky\ CPT, which we denote by \(\Phi_{X|\U},\) induces a conditional distribution \(\pr(X|\U)\) where:
\begin{equation} \label{eqn:model-def}
  \pr(X \mid \u) = 
\begin{cases} 
    \Theta_{X \mid \Delta_1} & \mbox{if } \u \models \Delta_1 \\
    ... & ... \\
    \Theta_{X \mid \Delta_k} & \mbox{if } \u \models \Delta_k 
\end{cases}.
\end{equation}
Since the contexts \(\Delta_i\) are mutually-exclusive and exhaustive, this conditional distribution is well-defined.
A context \(\Delta_i\) can be specified explicitly using a propositional sentence, like in a conditional PSDD.  It can also be specified using, for example, a neural network.

Consider the following \funky\ CPTs for a variable \(X\) with three parents \(U_1, U_2, U_3\):
\begin{center}
\small
\begin{tabular}{c|c}
\(\Theta_{X|\Delta_i}\) & \(\Delta_i\) \\ \hline
\((0.9,0.1)\) & if \(U_1 \oplus U_2 \oplus U_3 \equiv 1\) \\
\((0.1,0.9)\) & if \(U_1 \oplus U_2 \oplus U_3 \equiv 0\).
\end{tabular}
\quad
\begin{tabular}{c|c}
\(\Theta_{X|\Delta_i}\) & \(\Delta_i\) \\ \hline
\((0.9,0.1)\) & if \(f(u_1,u_2,u_3) \le T\) \\
\((0.1,0.9)\) & if \(f(u_1,u_2,u_3) > T\).
\end{tabular}
\end{center}
Here, all variables are binary (0/1).
The left \funky\ CPT has two contexts: (1) the parity of the parents is odd and the distribution over \(X\) is \((0.9,0.1)\), and (2) the parity is even and the distribution over \(X\) is \((0.1,0.9)\).  The right \funky\ CPT thresholds the output of a neural network \(f\) to define its contexts: if the output of the neural network \(f\) on inputs \(u_1,u_2,u_3\) is at or below our threshold, we are in the first context; otherwise we are in the second context.

\section{Learning} \label{sec:learn}

In this section, we show how to (1) learn the parameters of a \funky\ CPT when the contexts are known, and (2) how to learn the contexts of a \funky\ CPT, using parameter learning as a sub-routine.

\subsection{Learning the Parameters}

Given a dataset \(\data\), the log likelihood of a set of Bayesian network parameters \(\Theta\) is
\begin{align*}
LL(\data \mid \Theta) 
%& = \sum_{i=1}^N \log \pr(\d_i) 
%  = \sum_{i=1}^N \log \prod_{X\U} \pr(x_i|\u_i) 
%  = \sum_{i=1}^N \sum_{X\U}  \log \pr(x_i|\u_i) \\
%& = \sum_{X\U} \sum_{x\u} \pr_\data(x\u) \log \pr(x|\u) \\
& = \textstyle \sum_{X\U} CLL(\data_{X\U} \mid \Theta_{X|\U})
\end{align*}
which is the sum of the conditional log likelihoods of the CPTs \(\Theta_{X|\U}\) given the datasets \(\data_{\X\U}\) projected onto the families \(X\U\).  The local conditional log likelihoods is given by
\begin{align*}
CLL(\data_{X\U} \mid \Theta_{X|\U})
& = \textstyle \sum_{i=1}^N \log \theta_{x_i|\u_i} %\label{eqn:cll}
%& = \sum_{i=1}^N \log \pr_\theta(x_i \mid \u_i) \label{eqn:cll}
%  = \sum_{x\u} \pr_\data(x\u) \log \theta_{x|\u}
\end{align*}
which we can optimize independently.  Let \(\data\#(\y)\) denote the number of instances in the dataset \(\data\) compatible with the partial instantiation \(\y\).  The parameters \(\theta^\star_{x|\u}\) that optimize the conditional log likelihood is given by
\(
\theta^\star_{x|\u} = \frac{\data\#(x\u)}{\data\#(\u)}
\)
which further represent the maximum-likelihood estimates; 
for more details, see, e.g., \cite{Darwiche09,KollerFriedman,MurphyBook}.

Analagous estimates can be obtained for a Bayesian network with \funky\ CPTs \(\Phi_{X|\U}\) with contexts \(\Delta_i\) and the corresponding CPT columns \(\Theta_{X|\Delta_i}.\)  Namely, we have the maximum-likelihood estimates
\(
\theta^\star_{x|\Delta_i} = \frac{\data\#(x,\Delta_i)}{\data\#(\Delta_i)}.
\)
That is, we count the number of instances compatible with both \(x\) and the context \(\Delta_i\), and then normalize by the total number of instances compatible with the context \(\Delta_i\).

Each maximum likelihood estimates \(\theta^\star_{x|\Delta_i}\) for a \funky\ CPT \(\Phi_{X|\U}\) can be computed using a single pass of the dataset \(\data\), and also proportional to the time it takes to test whether a given example \(\d\) from the dataset is compatible with \(\Delta_i\).  If \(\Delta_i\) is represented using a propositional formula, we can simply evaluate \(\Delta_i\) given the partial instantiation \(\u\) found by projecting \(\d\) onto the parents \(\U\).  If \(\Delta_i\) is represented implicitly using a neural network, then it suffices to evaluate the network given partial instantiation \(\u\) and then perform the corresponding threshold test.

\subsection{Learning the Contexts}

Next, we propose a simple algorithm to learn the contexts of a \funky\ CPT from a given dataset \(\data\).

As discussed in Section~\ref{sec:model}, a thresholded multi-layer perceptron (MLP) can be used to implicitly define a set of contexts that partition the space of parent instantiations \(\u\).  Our approach has two steps: (1) we first learn a functional CPT using an MLP, like the CPTs we discussed in Section~\ref{sec:functional-cpts}, and then (2) we iteratively learn thresholds on the output, which in turn partitions the input space.  As we showed in Theorem~\ref{thm:difficult-structured-prediction}, MPE inference using a functional representation of a CPT is in general intractable, like the one we shall learn in Step~(1).  As we shall discuss later in Section~\ref{sec:reason}, the \funky\ CPTs that we obtain from Step~(2) shall give us a way to approach this apparently intractability.

First, we learn a functional representation of the conditional distribution \(\pr(X \mid \U).\)  Here, we use an MLP, which we denote by \(f_x(\u),\) to estimate the conditional probabilities \(\pr(x \mid \u)\) for some distinguished state \(x\) of a variable \(X\); for simplicity, we assume \(X\) is binary (0/1).  
%\art{describe how to handle multi-valued case}
In particular, the MLP is trained using feature-label pairs \((\u,x)\) for all family instantiations \(x,\u\) that appear in the original dataset \(\data\).  In our experiments, we used cross entropy as a loss function.

Our next step is to obtain a \funky\ CPT \(\Phi_{X|\U}\) from the MLP \(f_x(\U)\) that we have just learned.  Our approach is based on learning a threshold on the output of our MLP, which in turn induces a partition of the input space.  By iteratively learning additional thresholds, we can further refine our partitioning.
Suppose for now that we learn a single threshold \(T\), which yields the following partition:
\(
\Delta_{\le T} = \{ \u \mid f_x(\u) \le T \}
\)
and
\(
\Delta_{> T}   = \{ \u \mid f_x(\u)  >  T \}
\)
where \(\Delta_{\le T} = \neg \Delta_{> T}\) relative to all parent instantiations \(\u\), i.e., we have a partitioning.

%There are many equivalent threshold that are equally good according to our optimization objective. The context sentence \(\Delta_T\) partitions examples in \(\data\) into two parts. Some training examples \((x_i, \y_i)\) have the \(g(x^* \mid \y)\) values, which is computed by contextual function, below the threshold, and some are not. As long as two thresholds result in the same partitions of the training data, two thresholds are equally good according to our objective. Consider the training examples shown in Figure~\ref{fig:training}. Threshold \(0.4\) and \(0.45\) result in the identical partitions of the training records. The partition separates records \(\y_3\) from the rest of the examples. Hence, the two thresholds will result in identical scores in Equation~\ref{eqn:cll}.

\begin{figure}
  \small
  \setlength{\tabcolsep}{2pt}
  \centering
  \begin{subfigure}[b]{.15\linewidth}
    \centering
    \begin{tabular}{l|c c c}
      &\(U_1\) & \(U_2\) & \(X\) \\ \hline
      \(\d_1\) & \(0\) & \(0\) & \(1\) \\
      \(\d_2\) & \(1\) & \(0\) & \(0\) \\
      \(\d_3\) & \(1\) & \(1\) & \(0\) \\
      \(\d_4\) & \(0\) & \(1\) & \(1\) \\
      \(\d_5\) & \(1\) & \(1\) & \(1\)
    \end{tabular}
    \caption{\label{tab:data1}}
  \end{subfigure}
  \quad\quad
  \begin{subfigure}[b]{.19\linewidth}
    \centering
    \begin{tabular}{l|c c c}
      &\(U_1\) & \(U_2\) & \(f\) \\ \hline
      \(\d_2\) & \(1\) & \(0\) & \(\sigma(-2)\) \\
      \(\d_4\) & \(0\) & \(1\) & \(\sigma(-1)\) \\
      \(\d_3\) & \(1\) & \(1\) & \(\sigma(1)\) \\
      \(\d_5\) & \(1\) & \(1\) & \(\sigma(1)\) \\
      \(\d_1\) & \(0\) & \(0\) & \(\sigma(2)\)
    \end{tabular}
    \caption{\label{tab:data2}}
  \end{subfigure}
  \quad\quad
  \begin{subfigure}[b]{0.51\linewidth}
    \begin{tabular}{r|cc}
      \(T\) & \(\Delta_{\leq T}\) & \(\Delta_{> T}\) \\ \hline
      \(-\infty\) & \(\{\}\) & \(\{\d_2, \d_4, \d_3, \d_5, \d_1\}\) \\
      \(\sigma(-2)\) & \(\{\d_2\}\) & \(\{\d_4, \d_3, \d_5, \d_1\}\) \\
      \(\sigma(-1)\) & \(\{\d_2, \d_4\}\) & \(\{\d_3, \d_5, \d_1\}\) \\
      \(\sigma(1)\) & \(\{\d_2, \d_4, \d_3, \d_5\}\) & \(\{\d_1\}\) \\
      \(\sigma(2)\) & \(\{\d_2, \d_4, \d_3, \d_5, \d_1\}\) & \(\{\}\)
    \end{tabular}
    \caption{\label{fig:delta-t2}}
  \end{subfigure}
  \caption{(a) a dataset, (b) sorted by MLP output, (c) different thresholds and the resulting partitions. \label{fig:example}}
\end{figure}

Consider Figure~\ref{fig:example}, which highlights a simple example.  In Figure~\ref{tab:data1}, we have a small dataset.  Suppose that we learn the following MLP \(f\) from this dataset:
\(
f(u_1,u_2) = \sigma(6 u_1 u_2 - 4 u_1 - 3 u_2 + 2).
\)
In Figure~\ref{tab:data2}, we have sorted this dataset by the value of \(f(\u)\); remember that the sigmoid function \(\sigma\) is a monotonic non-decreasing function, i.e., \(\sigma(x) \le \sigma(y)\) iff \(x \le y\).  Note that for any two consecutive output values \(f_i\) and \(f_j\) in the sorted list of Figure~\ref{tab:data2}, any chosen threshold \(T \in (f_i,f_j]\) will result in the same partitioning.  For example a threshold \(\eql(T,0)\) results in the same partition as threshold \(\eql(T,0.9)\), yielding the sets \(\Delta_{\le T} = \{\d_2,\d_4\}\) and \(\Delta_{> T} = \{\d_3,\d_5,\d_1\}\) (note that \(\d_3\) and \(\d_5\) represent the same parent instantiation \(\eql(u_1,1),\eql(u_2,1)\)).

For each threshold \(T\), we can learn the resulting parameters \(\theta_{X|\Delta_{\le T}}\) and \(\theta_{X|\Delta_{> T}}\) using a single pass over the dataset, and then compute the resulting conditional log likelihood.  If \(N\) is the size of the dataset \(\data\), then it suffices to check \(N\) possible threshold values, plus one additional threshold \(\eql(T,-\infty)\) that ensures that \(\Delta_{\le T}\) is empty, and that \(\Delta_{> T}\) contains all of the examples.  We then simply pick the single threshold that maximizes the conditional log likelihood.  Finally, one can amortize the complexity of computing the conditional log likelihoods for all possible thresholds, hence requiring a single pass over the dataset \(\data\) overall.  
%This procedure is given in Algorithm~\ref{alg:context-learning}.

We can refine the partition further by recursing on each partition, and finding an additional threshold within each partition, using the same algorithm we described above.  We can continue to recurse and refine our partition, until validation likelihood falls or does not improve enough.

\section{Reasoning} \label{sec:reason}

In general, if we use a purely functional representation of a CPT, then inference becomes intractable, as given by Theorem~\ref{thm:difficult-structured-prediction}.  Alternatively, we seek next to obtain tractable \funky\ CPTs, first using recently proposed analytic tools from the domain of explainable AI (XAI).

\subsection{Marginal Inference via Knowledge Compilation}

Recently, in the domain of XAI, \cite{ChoiShiShihDarwiche19} showed how a binary neural network (BNN) can be formally analyzed and verified using symbolic tools from the domain of Knowledge Compilation \cite{darwicheJAIR02}.  A BNN is a neural network with binary inputs and a binary output.  Such a neural network represents a Boolean function.  Consider for example a linear classifier \(f\): \(1.15 \cdot U_1 + 0.95 \cdot U_2 - 1.05 \cdot U_3 \ge 0.52\).  Here, \(U_1,U_2,U_3\) are binary (0/1) inputs, and the classifier outputs 1 if this threshold test passes and it outputs 0 otherwise.  We can enumerate all possible inputs and record the classifier output \(f(u_1,u_2,u_3)\), leading to the following truth table:
\begin{center}
\begin{tabular}{ccc|c}
\(U_1\) & \(U_2\) & \(U_3\) & \(f\) \\ \hline
0 & 0 & 0 & 0\\
0 & 0 & 1 & 0\\
0 & 1 & 0 & 1\\
0 & 1 & 1 & 0
\end{tabular}
\quad
\begin{tabular}{ccc|c}
\(U_1\) & \(U_2\) & \(U_3\) & \(f\) \\ \hline
1 & 0 & 0 & 1\\
1 & 0 & 1 & 0\\
1 & 1 & 0 & 1\\
1 & 1 & 1 & 1
\end{tabular}
\end{center}
The original numerical linear classifier is thus equivalent to the propositional sentence \([\neg U_3 \wedge (U_1 \vee U_2)] \vee [U_3 \wedge U_1 \wedge U_2]\).  Previously, \citep{chanUAI03} showed how to extract the Boolean function of a given linear classifier, which includes neurons with step activations as a special case.  More recently, \cite{ChoiShiShihDarwiche19} showed that one can compose the Boolean functions of binary neurons and aggregate them to obtain the Boolean function of a binary neural network.  By compiling this Boolean function into a tractable logical representation, such as an Ordered Binary Decision Diagram (OBDD) or as a Sentential Decision Diagram (SDD), then certain queries and transformations can be performed in time that is polynomial in the size of the resulting circuit \cite{darwicheJAIR02,Darwiche11}.

Here, we use the algorithm proposed by \cite{ChoiShiShihDarwiche19}, to compile a binary neural network into an SDD circuit, in order to compile a \funky\ CPT into a Conditional PSDD.  First, if we threshold the output of an MLP with step-activations, then it corresponds to a binary neural network.  Hence, we can compile each \funky\ CPT context \(\Delta_i\) into an SDD.  Second, it is straightforward to compile a CPT column \(\Theta_{X|.}\) into a PSDD \citep{ShenCD18}.  It is then straightforward to aggregate all of the context SDDs and CPT column PSDDs into a single Conditional PSDD \citep{ShenCD18}.  If we obtain the CPTs of a Bayesian network as a Conditional PSDD, then we can employ the algorithms in \cite{ShenGCD19,ShenCD16}, in order to compute marginals in the Bayesian network.\footnote{Note that while multiplying two PSDDs is a tractable operation, multiplying \(n\) PSDDs may not be.}

\shrink{
Each context of a FCDT is described with a function, i.e. a MLP, with a set of thresholds. In this section, we explain how to convert each contexts into a tractable circuits. Ultimately, we can use these local tractable circuits to obtain a single tractable circuits representing the joint probability of the BN. This allows one to perform many queries efficiently, including most probable explanation, marginal queries.

Contextual functions are represented using MLPs. For the purpose of inference, we assume each hidden neuron is step activated, and the output neuron is sigmoid activated. During training, sigmoid activated neurons are preferred as one can apply SGD to optimize parameters of each neuron. To apply our inference procedure, one can replace each sigmoid activated hidden neurons to a step activated neurons without changing the weight. This is the same conversion procedure that is used in our experiments as well as in \cite{ChoiShiShihDarwiche19}.

We can apply the same algorithm from \cite{ChoiShiShihDarwiche19} to obtain a d-DNNF that describes parent configurations of each context. After we have compiled the contexts of a FCDT model into a tractable d-DNNF, one can utilize the same method that is described in \cite{ShenGCD19} to obtain a single d-DNNF that models the joint probability of the BN.
}

\subsection{MPE Inference via Mixed-Integer Linear Programming}

Consider the most probable explanation (MPE) query in a Bayesian network:
\(
\argmax_{\x \sim \e} \pr(\x),
\)
where \(\x\) is a complete instantiation of the network variables, \(\e\) is the observed evidence, and \(\sim\) denotes compatiability between \(\x\) and \(\e\) (they set common variables to the same values).  Computing the MPE is an NP-complete problem \cite{Shimony94}.  Theorem~\ref{thm:difficult-structured-prediction} shows that MPE is still NP-complete with independent parents \(\U\) and a common observed child \(X\) with a functional CPT.

Given a \funky\ CPT, we can apply a mixed-integer linear programming (MILP) solver to the task of solving an MPE query, i.e., to compute \(\argmax_{\u} \pr(\u,x)\) where \(\pr(X|\U)\) is represented with a \funky\ CPT.  First, the log of the MPE is a linear function of the log parameters of the network parameters, which we use as the objective of the MILP.  Using a \funky\ CPT for a binary variable \(X\), an observation \(x\) effectively adds another term to the objective function, which depends on the context implied by the input \(\u\).  This can be incorporated by the MILP after observing that an MLP with step activations can be reduced to an MILP, as in \cite{NarodytskaKRSW18,Griva08Linear}.  

\section{Experiments} \label{sec:experiments}

In this section, we empirically evaluate the \funky\ CPT representation that we proposed in Section~\ref{sec:model}, as well as the learning algorithm that we proposed in Section~\ref{sec:learn}.  In particular, we evaluate it in terms of our effectiveness at learning conditional distributions, in comparison to other functional and context-specific representations.  We shall subsequently evaluate the reasoning algorithms proposed in Section~\ref{sec:reason}, via a case study in Section~\ref{sec:case-study}.

We evaluate two sets of benchmarks, one synthetic, and one real-world.  We consider two baselines: (1) a functional CPT representation, using a multi-layer perceptron (MLP), and (2) a context-specific CPT representation, namely a tree CPT, which was learned using \idthree.  We compare each representation based on their (negated) conditional log likelihood (CLL); lower is better.
When learning any CPT column, we further use Laplace (add-one) smoothing.

We trained an MLP \(f_x(\U)\) to predict the value of variable \(X\) given an instantiation of the parents \(\U\).  We used a single hidden layer of \(16\) neurons with ReLU activations, whose parameters were learned with cross-entropy loss.  The MLP was trained using the Adam optimizer in \textsc{TensorFlow}.
We trained a tree CPT using the \idthree\ algorithm of the toolkit \textsc{scikit-learn}.  In our experiments, we learned decision trees of gradually increasing complexity (measured by counting decision tree leaves), by gradually increasing the bound on tree depth, which is a parameter of the \idthree\ algorithm.
%In particular, at eath step, the \idthree picks a parent variable to expand each leaf. The parent variable is picked that optimize the training CLL. 
%
%After the tree structure is learned through \idthree, we estimate the leaf probability using the same Equation, while applying the \(1\)-laplacian smoothing.

Finally, to obtain a \funky\ CPT, we use the learning algorithm described in Section~\ref{sec:learn}, using the MLP that we trained above.  In our experiments, we also gradually increased the number of contexts created.  With \(k\) thresholds created, we create \(k+1\) contexts, which we compare with the number of contexts found by the tree CPT (i.e., the number of decision tree leaves).

%After the MLP is obtained, we apply the threshold learning algorithm to learn a threshold. The threshold creates two contexts for the \funky, \(\Delta_{\leq}\) and \(\Delta_>.\) To obtain more contexts, we can apply the same algorithm again to find a threshold to further partition the context \(\Delta_{\leq}\) or \(\Delta_>.\) By recursively applying our algorithm, we are able to obtain a \funky model with arbitrary number of contexts. Further, the leaf probability of each context is estimated using Equation, while we apply a 1-laplacian smoothing.

\paragraph{Synthetic Benchmark.}  In our synthetic experiment, we simulated training data from a conditional distribution exhibiting many context-specific independencies.  That is, we simulated data where the value of variable \(X\) depends only on the \emph{cardinality} of its parents \(\U\), where \(X\) and \(\U\) are binary (0/1).  If the fraction of parents set to 1 is at most \(\frac{1}{k}\), then \(\pr(x|\u) = 0.05\); otherwise 
%the fraction of parents is strictly greater than \(\frac{1}{k}\) and 
\(\pr(x|\u) = 0.95\).  In our experiments, we simulated parent instantations \(\u\) such that the two different contexts (\(\le \frac{1}{k}\) and \(> \frac{1}{k}\)) had the same probability of being generated.

\begin{figure}[tb]
 \centering
 \includegraphics[width=.32\linewidth]{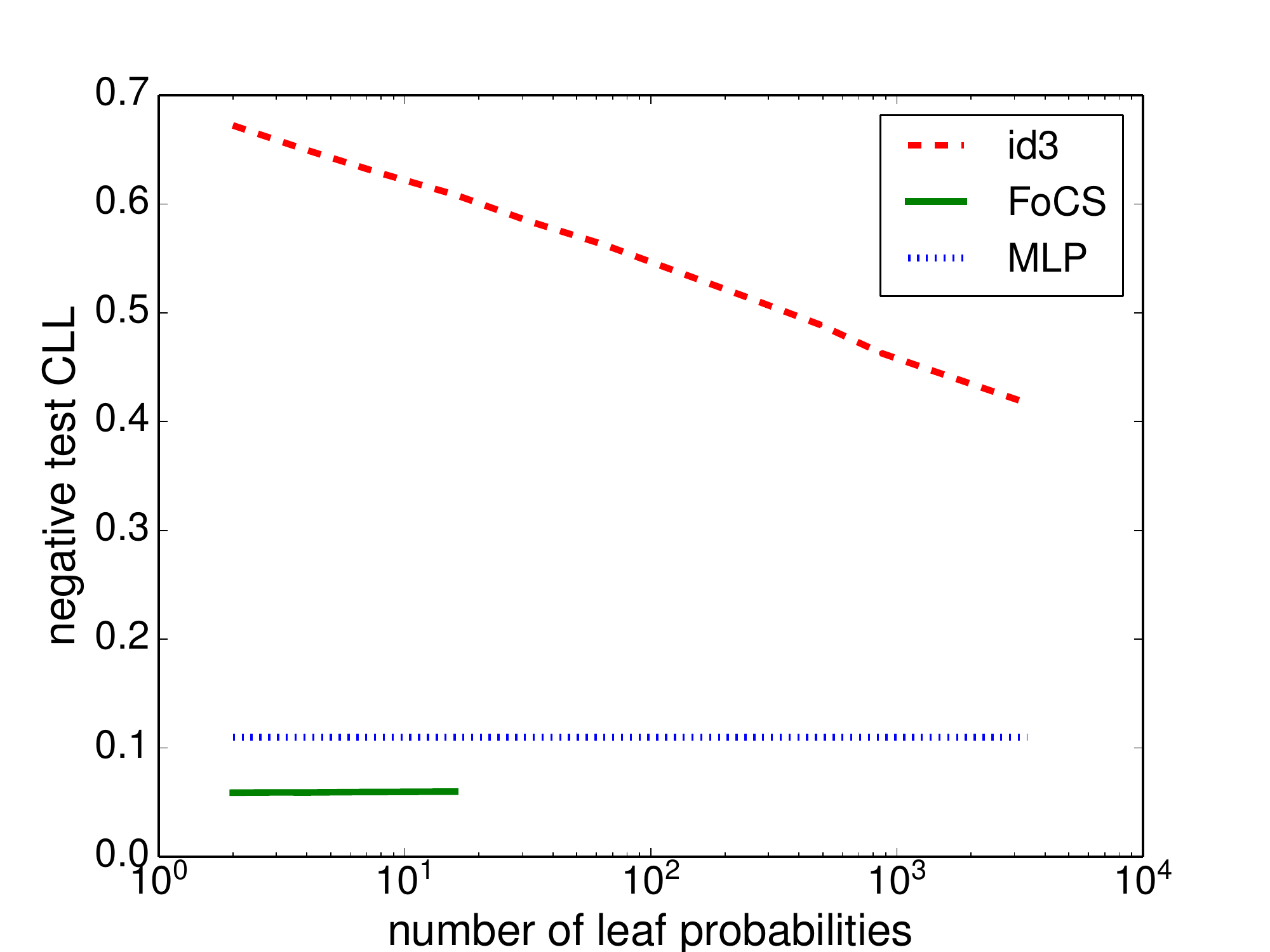}  
 \includegraphics[width=.32\linewidth]{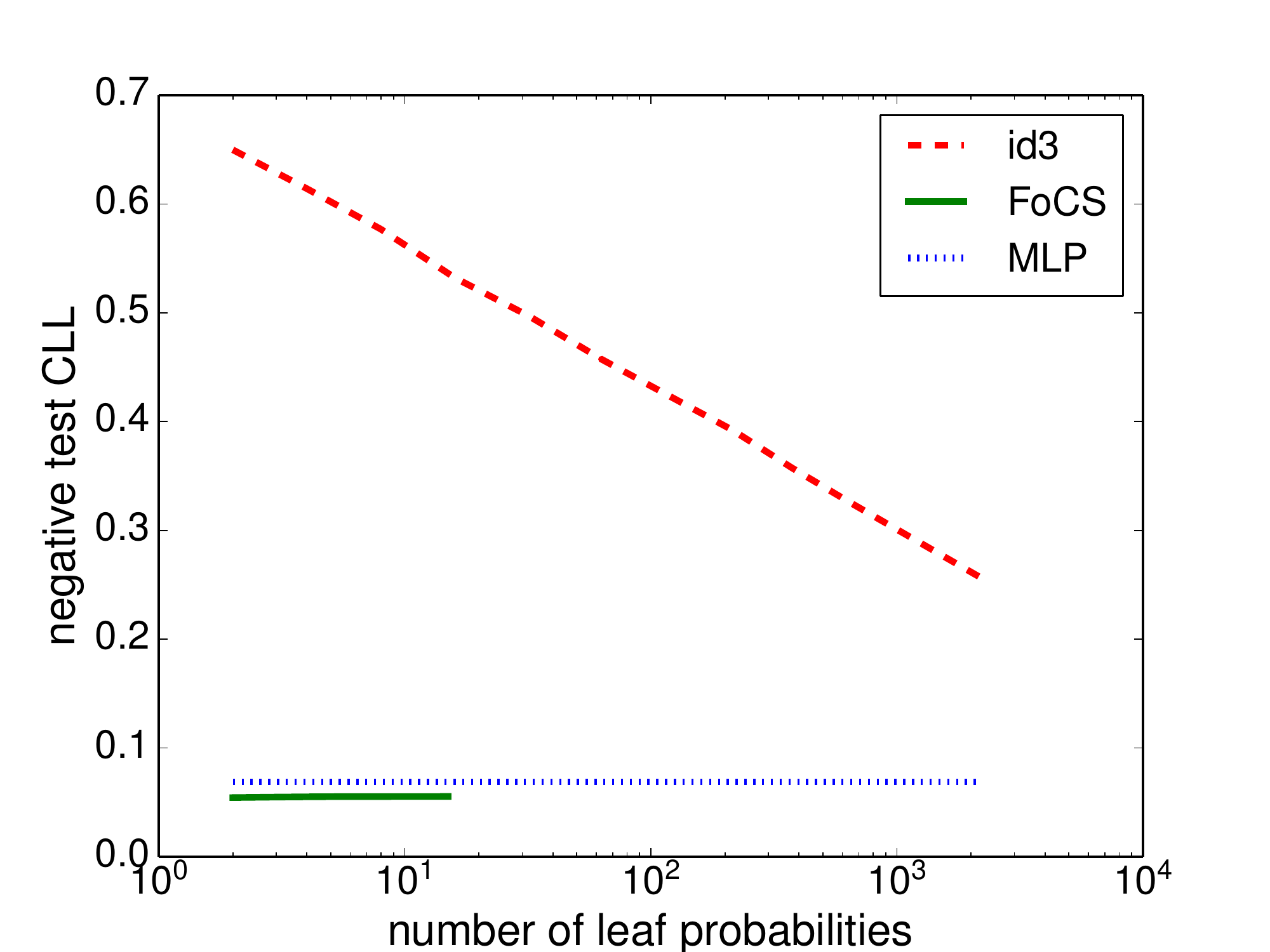} 
 \includegraphics[width=.32\linewidth]{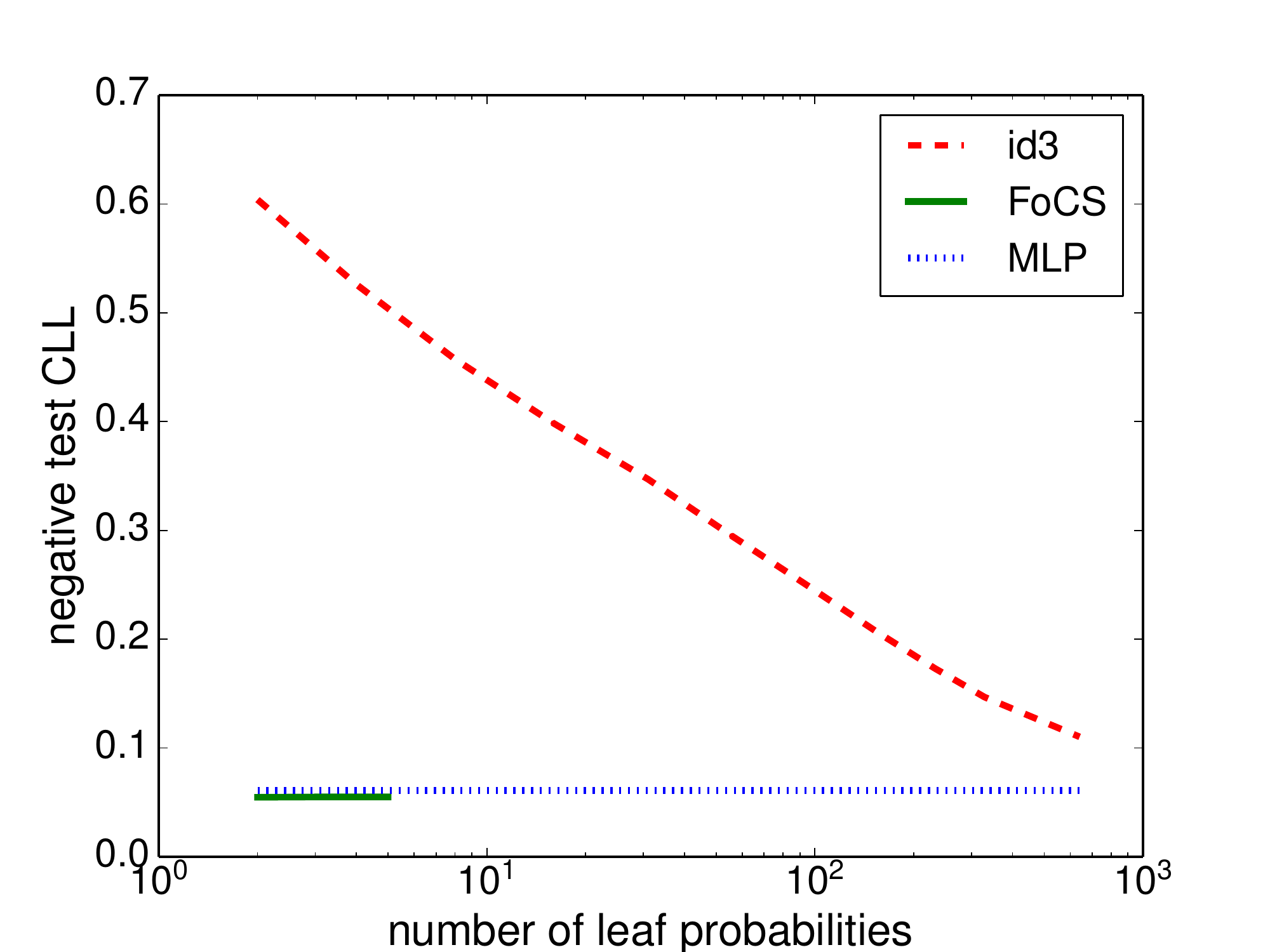}
\caption{Number of contexts vs. CLL, for \(k \in \{2,4,8\}\) from left to right (synthetic benchmark).
\label{fig:card}}
\end{figure}

In Figure~\ref{fig:card}, we plot results for \(k \in \{2,4,8\}\).  On the \(x\)-axis we increase the number of contexts for the tree CPT (by increasing the bound on depth) and for the \funky\ CPT (by adding more thresholds); the MLP is a functional CPT without any explicit CSIs, and hence is a flat line on each plot.  We make the following observations in Figure~\ref{fig:card}: (1) as we increase the number of contexts of the tree CPT (\idthree), the better the CLL, (2) the MLP and the \funky\ CPT perform similarly, and both obtain better CLLs than the tree CPT, and (3) the \funky\ CPT obtains a good CLL using only a small number of contexts, and obtains a better CLL than the MLP that it was created from.

It is well-known that decision trees cannot succinctly represent certain (Boolean) functions.  For example, a decision tree must be complete, using \(2^n\) leaves, to represent the parity function over \(n\) variables.  This is also the case for cardinality constraints over \(n\) variables, which have less succinct decision trees for \(k=2\) and succinct decision trees for \(k=1\) or \(k=n\).  We see this pattern as well in Figure~\ref{fig:card}, as the performance of \idthree\ more closely approaches that of MLP and our \funky\ CPT.

%\cite{Swersky12,ShenChoiDarwiche17}

Compared to the MLP, our \funky\ model estimates much fewer parameters---once we are given \(k\) contexts \(\Delta_i\), then we simply need to estimate the \(k\) corresponding CPT columns \(\Theta_{X|\Delta_i}\).  The fact that our learning appears to converge almost immediately, suggests that our learning algorithm is indeed learning the context-sensitivies inherent in the cardinality constrained data that we simulated.  In contrast, the MLP does not search for CSIs.  Hence, this explains the ability of our \funky\ model to obtain better CLLs than the MLP that our \funky\ model was based on.

%\jason{describes about the tree CPT}

\paragraph{Real-World Benchmark.} Next, we consider a real-world dataset: MNIST digits.  This dataset is composed of \(28 \times 28\) pixel grayscale images, which we binarized to black-and-white.  We consider one-vs-all classification, where the parents \(\U\) represent the input image, and the child \(X\) represents whether the input is of a particular digit \(d\) (\(\eql(X,\true)\)) or some other digit from \(0\) to \(9\) (\(\eql(X,\false)\)).

\begin{figure}[tb]
 \centering
 \includegraphics[width=.32\linewidth]{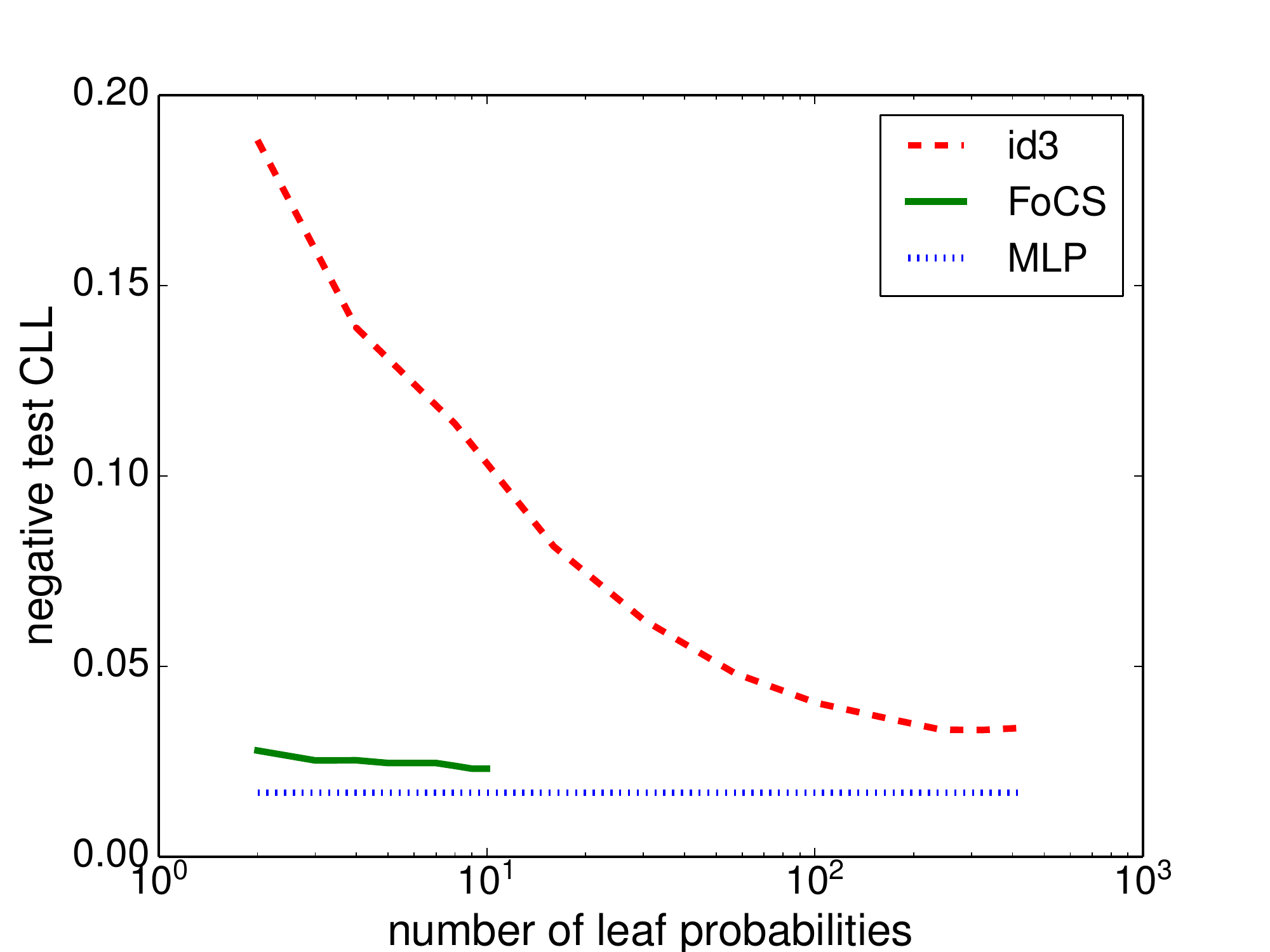}  
 \includegraphics[width=.32\linewidth]{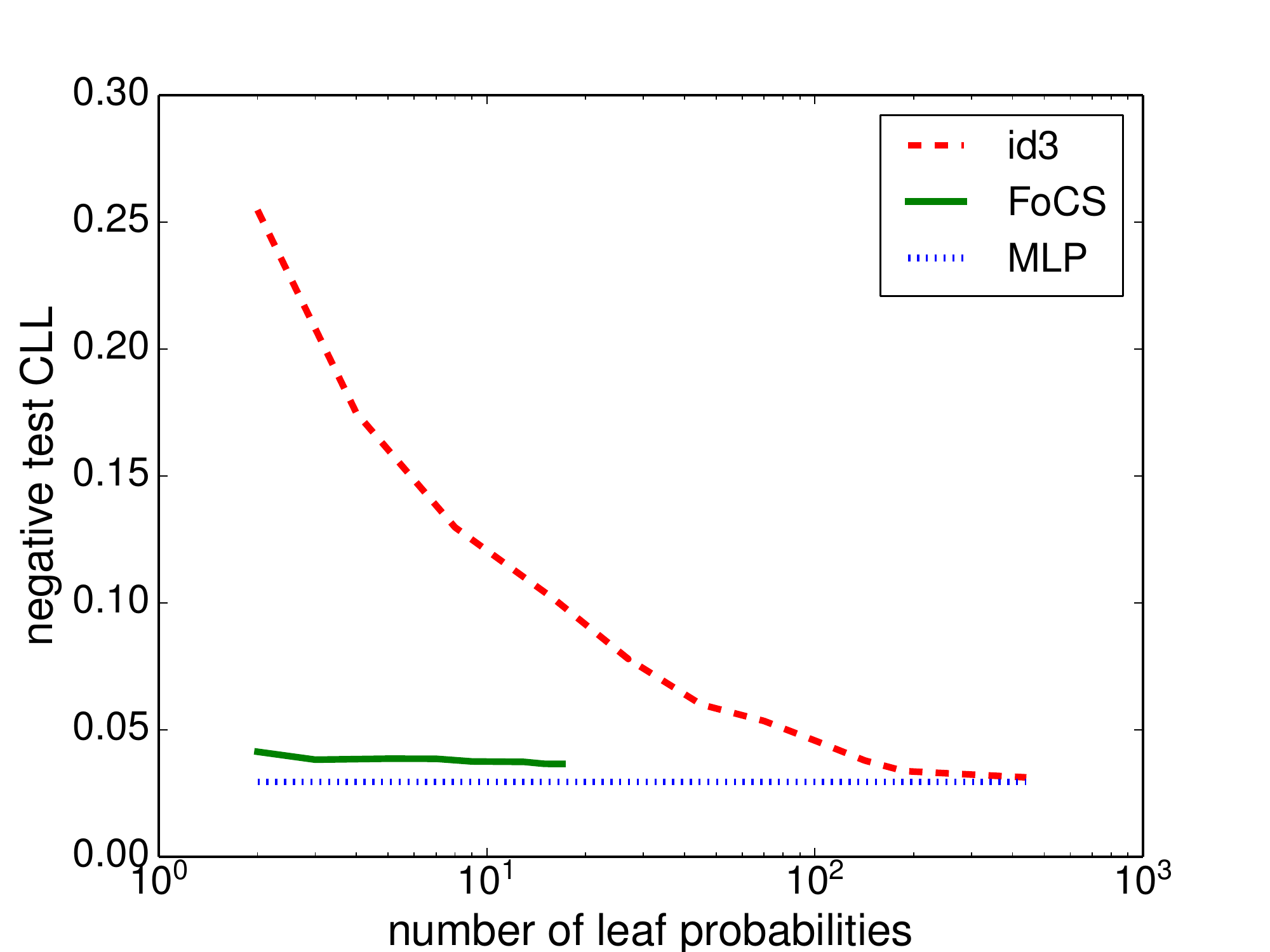} 
 \includegraphics[width=.32\linewidth]{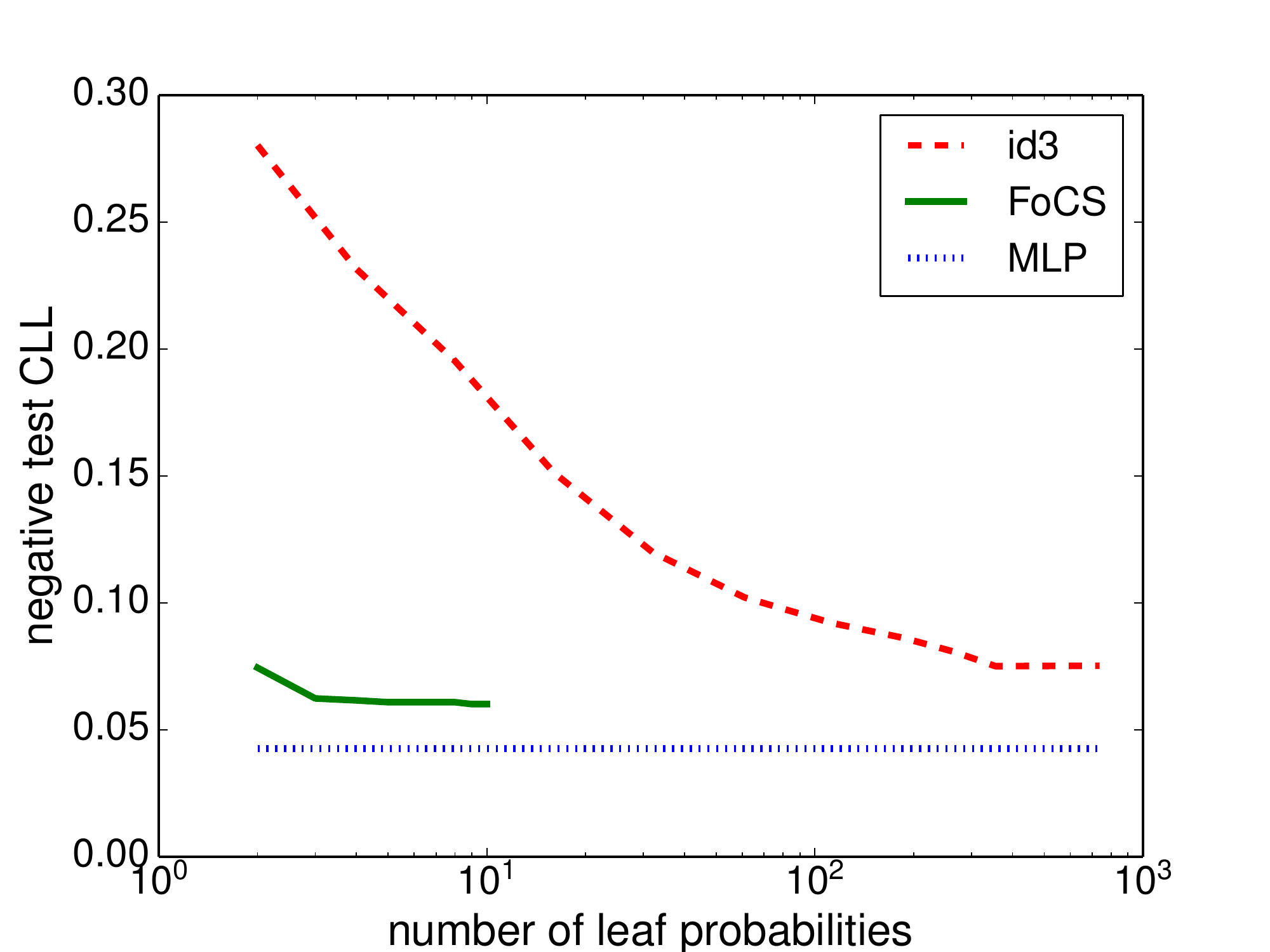}
\caption{Number of contexts vs. CLL, for digits \(d \in \{0,1,2\}\) from left-to-right (MNIST).
\label{fig:digits}}
\end{figure}

Figure~\ref{fig:digits} highlights the result.  Our \funky\ model consistently estimates the conditional distribution more accurately using fewer contexts compared to the tree CPT model.  This provides strong evidence that exploiting the structure from a learned functional model is more efficient than searching for the structure of contexts by variable-splitting as done when learning a decision tree.  Again, given that our \funky\ model appears to converge relatively quickly, this suggests that our learning algorithm is able to learn the CSIs from data (a CSI may represent here a partial instantiation of the input pixels that will almost guarantee the classification of a digit).  However, a small number of contexts may not be enough to obtain the performance of an MLP.  Note that there may be only a limited number of contexts that our learning algorithm can discover, if the training data is not diverse enough, hence why the curve for our \funky\ model stops earlier than that for the tree CPT.\footnote{For example, if the MLP is trained to the point where it obtains \(100\%\) confidence in most of the training examples (which is not unlikely for datasets such as MNIST), then we would not be able to split any of the resulting contexts).}

\section{Case Study: Learning to Decode} \label{sec:case-study}

In this section, we show through a simple case study how \funky\ CPTs allow us to reason about and learn from complex processes.  Our case study is done in the context of channel coding, where our goal is to encode data in a way that allows us to detect and correct for any errors that may occur after transmission through a noisy channel.  In subsequent experiments, we show, using our proposed learning algorithm, how one can learn to decode encoded messages, without knowing the original code that was used to encode them!

\subsection{Channel Coding: A Brief Introduction}

Consider the following problem.  Say you have a message represented using \(n\) bits \(U_0,\ldots,U_{n-1}\) that can be either 0 or 1.  We want to transmit this message across a noisy channel, where there is a chance that each bit \(U_i\) might be corrupted by noise (say flipped from 0 to 1, or from 1 to 0).  To improve the reliability of this process we can send additional bits, say \(m\) of them \(X_0,\ldots,X_{m-1}\).  We refer to the original bits \(\u\) as the \emph{message} and the redundant bits \(\x\) as the \emph{encoding} (or alternatively the \emph{channel input}).  We further refer to the encoding process as the \emph{code}.  The \emph{channel output} are the bits \(\y\) received from the noisy channel.  Finally, there is a \emph{decoding} process that attempts to detect and correct any errors in the channel output.

One simple example of a code, is the repetition code, which sends \(k\) additional copies of the message across the noisy channel (say 3 copies total).  At the channel output, one detects an error if any of the 3 copies reports a discrepency among the corresponding bits.  One can attempt to correct for the error by taking a majority vote.
Repetition codes are among the simplest type of error-correcting code.  More sophisticated codes include \emph{turbo codes} and \emph{low density parity check codes}, whose decoders were shown to be instances of loopy belief propagation in a Bayesian network; see \citep{FreyM97} for a short perspective.  Another common type of code uses parity checks among randomly selected sets of message bits, as redundant bits in the encoding.

\subsection{Experiments}

\begin{figure}[t]
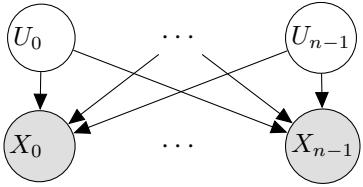

  \centering
  \tikz{
    \node[] (U_m) {$\cdots$};
    \node[latent, left=of U_m] (U_1) {$U_{0\hphantom{+1}}$};
    \node[latent, right=of U_m] (U_n) {$U_{n-1}$};
    \node[ below=of U_m] (X_mid) {$\cdots$};
    \node[obs, left=of X_mid] (X_1) {$X_{0\hphantom{+1}}$};
    \node[obs, right=of X_mid] (X_m) {$X_{n-1}$};
    \edge[->] {U_1} {X_1};
    \edge[->] {U_1} {X_m};
    \edge[->] {U_n} {X_1};
    \edge[->] {U_n} {X_m};
    \edge[->] {U_m} {X_1};
    \edge[->] {U_m} {X_m};
  }
  \caption{The Bayesian network modeling the encoding process. \label{fig:structured-prediction}}
\end{figure}

We can model a message encoder using a Bayesian network like the one in Figure~\ref{fig:structured-prediction}.  Root nodes \(U_i\) represent the bits (0/1) to be encoded, and the leaf nodes \(X_i\) represent the encoded bits (0/1) that are to be transmitted across a noisy channel.  For simplicity, we assume that both the message and the encoding are composed of \(n\) bits.  We assume the message bits \(U_i\) are marginally independent.  Each of the encoded bits \(X_i\) can in general depend on any of the message bits \(U_i\), depending on the particular code being used.  At the same time, we can model the noisy channel, which may flip a bit from 0 to 1 or from 1 to 0, with some probability.  The traditional reasoning task would be, given a code, i.e., the conditional distributions \(\pr(X_i \mid \U)\), and a message \(\x\) received over the noisy channel, to find the most likely message \(\u\) that was originally encoded.

Consider for example a simple code where we record the parity of every pair of adjacent message bits.  Such an encoding, can be represented using the following CPT:
\[
\pr(\eql(X_i, 1) \mid \U) = \begin{cases} 100 \% &  \mbox{if } U_i \oplus U_{i+1 \bmod n} = 1\\
    0 \% & \mbox{if } U_i \oplus U_{i+1 \bmod n} = 0 \end{cases}.
\]
We can further model the noise in the channel with the following modified CPT.
\[
\pr(\eql(X_i, 1) \mid \U) = \begin{cases} 95 \% &  \mbox{if } U_i \oplus U_{i+1 \bmod n} = 1\\
    5 \% & \mbox{if } U_i \oplus U_{i+1 \bmod n} = 0 \end{cases}.
\]
Given a set of message/encoding pairs \((\u,\x)\) we can try to learn the code used to encode the messages, i.e., learn the conditional distributions \(\pr(X_i \mid \U)\).  We represent each conditional distribution using a \funky\ CPT, as a tabular representation would be intractable for this type of problem: the table would have a number of entries that is exponential in \(n\), and we would need at least as much data to learn the parameters.

We learn a \funky\ CPT as described in Section~\ref{sec:learn}, starting with an MLP with a single hidden layer containing \(8\) neurons with sigmoid activations.  We convert the sigmoid activations to step activations for the purposes of reducing it to MILP, in order to perform MPE inference as described in Section~\ref{sec:reason}.  From the MLP, we obtain a \funky\ CPT by learning one threshold, which yields \(2\) different contexts.

Once we have learned a \funky\ model from data, we can then try to decode an encoded message.  That is, given an encoded message \(\x\), we can find the most likely original encoding via: \(\argmax_{\u} \pr(\u \mid \x)\), which is an MPE query.  We used the MILP solver \textsc{Gurobi} \cite{gurobi} with the \textsc{cvxpy} optimizer \cite{agrawal2018rewriting} to solve these MILP problems, in our experiments, which we present next.

To obtain a training and testing set of messages \(\u\) we sampled bits at random with \(\pr(u_i) = 0.8\).  To obtain a set of encoded messages \(\x\), we used a code where each encoded bit took the parity of three consecutive bits (there are \(n\) such encoded bits).  We assume that the channel has a \(5\%\) chance of flipping an transmitted bit.  We simulated datasets of size \(2^{14} = 16,384,\) and performed \(5\)-fold cross validation.
The following table summarizes our results.
\begin{center}
  \begin{tabular} {l c c c r} \toprule
    \(n\) & word accuracy & bit accuracy & Hamming error & time (s) \\ \midrule
    \(10\) & \(0.750 \pm 0.003\) & \(0.902 \pm 0.002\)& \(0.974 \pm 0.021\) & \(0.247 \pm 0.001\) \\
    \(15\) & \(0.651 \pm 0.005\) & \(0.900 \pm 0.002\) & \(1.493 \pm 0.044\) & \(0.469 \pm 0.004\) \\
    \(20\) & \(0.578 \pm 0.005\) & \(0.905 \pm 0.001\) & \(1.886 \pm 0.037\) & \(1.047 \pm 0.036\) \\
    \(25\) & \(0.493 \pm 0.007\) & \(0.905 \pm 0.001\) & \(2.371 \pm 0.043\) & \(11.382 \pm 0.549\) \\
    \(30\) & \(0.414 \pm 0.006\) & \(0.901 \pm 0.003\) & \(2.963 \pm 0.099\) & \(140.190 \pm 11.539\)\\
    \bottomrule
  \end{tabular}
\end{center}
From top-to-bottom, each row represents increasing message sizes.  We report word accuracy (the percentage of instances where the original message was successfuly decoded from the encoding without error), bit accuracy (the percentage of bits that were decoded without error), Hamming error (the average number of incorrect bits in a decoded message), and time (in seconds).

We make a few observations.  Bit accuracy remains consistent around \(90\%\), for all message sizes \(n\).  Word accuracy falls, as expected, since it becomes more difficult to decode the entire message without error, the longer the message gets.  Note that a \(41.4\%\) for \(n=30\) is quite good compared to the expected word accuracy one would have obtained by composing a message estimate from most-likely-bit estimates at \(90\%\) accuracy, which would be \(0.9^{30} = 4.24\%\).  When we consider the hamming error, even if there were an error in the decoding, only a few bits were incorrect on average.  Finally, we see that inference time appears to grow exponentially as \(n\) grows.  This is also expected as decoding is in general an NP-hard problem.

%\jason{Using the Algorithm~\ref{sec:learn}, we are able to estimate the local probabilities in the Bayesian network in Figure~\ref{fig:structured-prediction}. After the BN is learned, we can further apply probabilistic reasoning to carefully handle the uncertainty in the problem. The combintation of the modeling and reasoning allows us to accurately predict labels jointly.}

%From the hamming distance, we can calculate the bit accuracy (the percentage of the time that each bit in the label is predicted accuratly) to be around \(90\%\). If each bit is were predicted independently without the model in Figure~\ref{fig:structured-prediction}, the word accuracy would have been only \(4\%\) instead of \(41\%\) to predict the original message of length \(30.\)

%\jason{Regarding complexity of reasoning, decoding the parity code, which is our ground truth, is NP-complete. As we expect to recover the truth coding function, decoding the learned model is expected to be equally hard. This reflects on the time column. The time grows expoentially with the number of bits in the original message. This computation bottleneck can be relaxed with future research on approximation methods that can be applied on models with context-specific independence.}

\section{Conclusion} \label{sec:conclusion}

We proposed here the \funky\ CPT model, for representing CSIs in conditional distributions.  We proposed an algorithm for learning the parameters as well as the contexts of \funky\ CPTs.  We showed how efficient inference can be enabled using \funky\ CPTs, by leveraging tools from knowledge compilation and optimization.  We highlighted some of the advantages of \funky\ CPTs compared to more traditional functional and context-sensitive CPT representations.  Finally, we provided a case study showing how \funky\ CPTs enable us to ``learn how to decode.''

\appendix

%\acks{We would like to acknowledge support for this project from the Random Science Foundation.}

\bibliography{bib/references}
\end{document}